\documentclass[10pt,twocolumn,letterpaper]{article}

\usepackage{cvpr}
\usepackage{times}
\usepackage{epsfig}
\usepackage{graphicx}
\usepackage{amsmath}
\usepackage{amssymb}

\usepackage{etoolbox}
\makeatletter
\patchcmd{\@makecaption}
  {\scshape}
  {}
  {}
  {}
\makeatother

\usepackage{bm}
\usepackage{xcolor}
\usepackage{graphics} 
\usepackage{epsfig} 
\usepackage{mathptmx} 
\usepackage{times} 
\usepackage{amsmath} 
\usepackage{amssymb}  
\usepackage{graphicx}
\usepackage{multirow, makecell}
\usepackage{array}
\usepackage{booktabs} 
\usepackage{comment}
\usepackage{commath}
\usepackage{subfig}
\usepackage{cite}
\usepackage[linesnumbered,ruled]{algorithm2e} 

\usepackage{ragged2e}
\newcolumntype{P}[1]{>{\RaggedRight\hspace{0pt}}p{#1}}
\newcolumntype{M}[1]{>{\centering\arraybackslash}m{#1}}

\newcommand{\beginsupplement}{%
        \setcounter{table}{0}
        \renewcommand{\thetable}{S\arabic{table}}%
        \setcounter{figure}{0}
        \renewcommand{\thefigure}{S\arabic{figure}}%
        \setcounter{section}{0}
        \renewcommand\thesection{\Alph{section}}
     }

\newenvironment{packed_enum}{
\begin{enumerate}
  \setlength{\itemsep}{1pt}
  \setlength{\parskip}{0pt}
  \setlength{\parsep}{0pt}
}{\end{enumerate}}

\usepackage[pagebackref=true,breaklinks=true,letterpaper=true,colorlinks,bookmarks=false]{hyperref}

\cvprfinalcopy 

\def\cvprPaperID{10268} 

\ifcvprfinal\fi
\begin{document}

\title{HCNAF: Hyper-Conditioned Neural Autoregressive Flow and its Application for Probabilistic Occupancy Map Forecasting}

\author{Geunseob (GS) Oh\\
University of Michigan\\
{\tt\small gsoh@umich.edu}
\and
Jean-S\'ebastien Valois\\
Uber ATG\\
{\tt\small jsvalois@uber.com}
}

\maketitle
\thispagestyle{empty}

\begin{abstract}
We introduce Hyper-Conditioned Neural Autoregressive Flow (HCNAF); a powerful universal distribution approximator designed to model arbitrarily complex conditional probability density functions. HCNAF consists of a neural-net based conditional autoregressive flow (AF) and a hyper-network that can take large conditions in non-autoregressive fashion and outputs the network parameters of the AF. Like other flow models, HCNAF performs \textit{exact} likelihood inference. We conduct a number of density estimation tasks on toy experiments and MNIST to demonstrate the effectiveness and attributes of HCNAF, including its generalization capability over unseen conditions and expressivity. Finally, we show that HCNAF scales up to complex high-dimensional prediction problems of the magnitude of self-driving and that HCNAF yields a state-of-the-art performance in a public self-driving dataset.
\end{abstract}

\vspace{-2pt}
\section{Introduction}
\label{sec:introduction}

Recent autoregressive flow (AF) models \cite{ref:2016_IAF, ref:2017_MAF, ref:2018_NAF, ref:2019_BNAF} have achieved state-of-the-art performances in density estimation tasks. They offer compelling properties such as \textit{exact} likelihood inference and expressivity. Of those, \cite{ref:2018_NAF, ref:2019_BNAF} successfully unified AF models and neural networks, and demonstrated an ability to capture complex multi-modal data distributions while universally approximating continuous probability distributions.

However, due to scalability limitation, existing neural AF models are ineffective at tackling problems with arbitrarily high-dimensional conditional terms. Scene prediction for autonomous driving is such a task where the benefits of AF models can be leveraged but where the contextual information (conditional terms) is too large (i.e. $C > \mathbb{R}^{1,000,000}$ due to using many multi-channel spatio-temporal maps). In contrast, the biggest experiment neural AF models reported is BSDS300 ($\mathbb{R}^{63}$) \cite{ref:2001_BSDS}. This may explain their limited use in common problems despite demonstrating excellent performance in density estimations. 

\begin{figure}[t]
    \centering
    \includegraphics[width=0.95\linewidth]{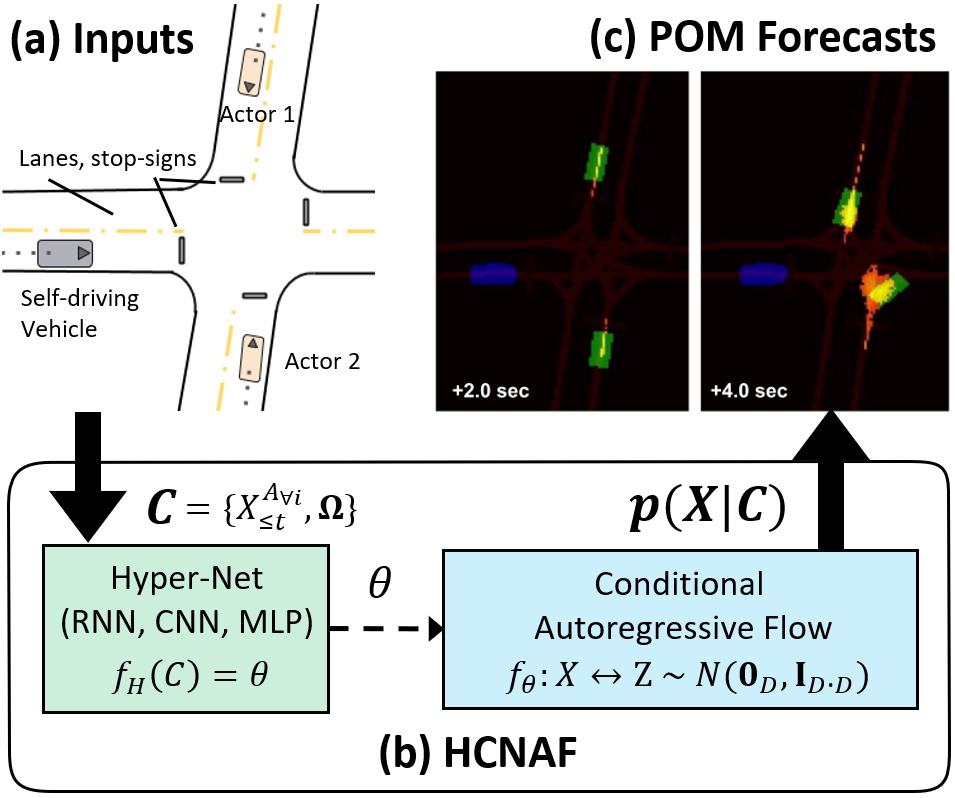}
    \caption{HCNAF used for probabilistic occupancy map (POM) forecasting, demonstrating the network's use of high-dimensional conditions ($C=\mathbb{R}^{>1mil}$). a) Inputs (conditions) are the spatio-temporal scene data. b) HCNAF consists of two neural-net based modules: a hyper-network $f_{H}$ and a conditional AF $f$. $f_{H}$ can take arbitrarily large inputs and produces the network parameters for $f$, which produces the conditional probability $p(X|C)$ precisely. c) Resulting POMs for agent vehicle centers at t=2 and t=4 secs.}
    \label{fig:hcnaf_overview}
\end{figure}

We propose a novel conditional density approximator called Hyper-Conditioned Neural Autoregressive Flow (HCNAF) to address the aforementioned limitation. HCNAF performs an \textit{exact} likelihood inference by precisely computing probability of complex target distributions $p_{model}(X|C) \approx p(X|C)$ with arbitrarily large $C$. By taking advantage of the design, HCNAF grants neural AFs the ability to tackle wider range of scientific problems; which is demonstrated by the autonomous driving prediction tasks.

Prediction tasks in autonomous driving involve transforming the history of high dimensional perception data up to the current time into a representation of how the environment will evolve \cite{ref:2018_FaF, ref:2018_intentnet, ref:2019_Multiple_Pred, ref:2017_Desire, ref:2019_PRECOG, ref:2019_Sophie, ref:2018_SGAN, ref:2019_Traphic, ref:2019_Intention, ref:2019_CoRL}. To be effective, advanced predictions models should exhibit the following properties:

\vspace{-5pt}
\begin{packed_enum}
    \item \textit{probabilistic}: reflecting future state uncertainties,
    \item \textit{multimodal}: reproducing the rich diversity of states, 
    \item \textit{context driven}: interactive \& contextual reasoning, and
    \item \textit{general}: capable of reasoning unseen inputs.
\end{packed_enum}

To incorporate the above requirements, we leverage HCNAF's powerful attributes such as the expressivity to model arbitrarily complex distributions and the generalization capability over unseen data. Furthermore, we opted for probabilistic occupancy maps (POMs) (see figure \ref{fig:hcnaf_overview}) over a more widely used trajectory-based prediction approach. As POM naturally encodes uncertainty, a POM represents all possible trajectories; thus removes the need to exhaustively sample trajectories like in trajectory-based methods.

Before presenting results on self-driving scenarios, we first introduce HCNAF and report results from a number of density estimation tasks to investigate HCNAF's expressivity and generalization capability over diverse conditions.

\vspace{-2pt}
\section{Background}
\label{sec:background}

Flow, or normalizing flow, is a type of deep generative models which are designed to learn data distribution via the principle of maximum likelihood \cite{ref:2016_NIPS_Tutorial_GM} so as to generate new data and/or estimate likelihood of a target distribution. 

Flow-based models construct an invertible function $f(z)=x$ between a latent variable $z$ and a random variable $x$, which allows the computation of exact likelihood of an unknown data distribution $p(x)$ using a known pdf $\pi(z)$ (e.g. normal distribution), via the change of variable theorem:

\vspace{-3pt}
\begin{equation}
\label{Eq:Change_of_Variables}
\begin{aligned}
p(x) = \pi(f^{-1}(x))\abs{det\frac{df^{-1}(x)}{dx}}.
\end{aligned}
\end{equation}

In addition, flow offers data generation capability by sampling latent variables $z \sim \pi()$ and passing it through $f$. As the accuracy of the approximation $f(z) = x$ increases, the modeled pdf $p_{model}(x)$ converges to the true $p(x)$ and the quality of the generated samples also improves.

In contrast to other classes of deep generative models (namely VAE\cite{ref:2013_VAE} and GAN\cite{ref:2014_GAN}), flow is an explicit density model and offers unique properties:

\vspace{-3pt}
\begin{enumerate}
    \item Computation of an \textbf{exact} probability, which is essential in the POM forecasting task. VAE infers $p(x)$ using a computable term; Evidence Lower BOund (ELBO). However, since the upper bound is unknown, it is unclear how well ELBO actually approximates $p(x)$ and how ELBO can be utilized for tasks that require exact inference. While GAN proved its power in generating high-quality samples for image generation and translation tasks\cite{ref:2017_image_translation_cGAN, ref:2018_starGAN}, obtaining the density estimation and/or probability computation for the generated samples is non-trivial. \vspace{-3pt}
    \item The expressivity of flow-based models allows the models to capture complex data distributions. A recently published AF model called \textit{Neural Autoregressive Flow (NAF)}\cite{ref:2018_NAF} unified earlier AF models including \cite{ref:2016_IAF, ref:2017_MAF} by generalizing their affine transformations to arbitrarily complex non-linear monotonic transformations. Conversely, the default VAE uses \textit{unimodal} Gaussians for the prior and the posterior distributions. In order to increase the expressivity of VAE, some have introduced more expressive priors \cite{ref:2017_VQVAE} and posteriors \cite{ref:2016_IAFVAE, ref:2018_Sylvester_NF} that leverage flow techniques.
\end{enumerate}

The class of invertible neural-net based autoregressive flows, including \textit{NAF} and \textit{BNAF}\cite{ref:2019_BNAF}, can approximate rich families of distributions, and was shown to universally approximate continuous pdfs. However, \textit{NAF} and \textit{BNAF} do not handle \textit{external} conditions (e.g. classes in the context of GAN vs cGAN\cite{ref:2014_cGAN}). That is, those models are designed to compute $p(x_{t})$ conditioned on previous inputs $x_{1:t-1}$ autoregressively to formulate $p(x_{t}|x_{1:t-1})$. This formulation is not suitable for taking arbitrary conditions other than the autoregressive ones. This limits the extension of \textit{NAF} to applications that work with conditional probabilities $p(X|C)$, such as the POM forecasting.


$MAF$ and $cMAF$ were proposed in \cite{ref:2017_MAF} to model affine flow transformations with and without additional external conditions. As shown in Equation \ref{Eq:flow_transformation_cMAF}, the transformation between $z_{d}$ and $x_{d}$ is affine and the influence of $C$ over the transformation relies on $\mu$, $\sigma$, and stacking multiple flows. These may limit the contributions of $C$ to the transformation. This explains the needs for a conditional autoregressive flow that does not have such expressivity bottleneck.

\vspace{-7pt}
\begin{equation}
\label{Eq:flow_transformation_cMAF}
cMAF: x_{d} = \mu(x_{1:d-1},C) + \sigma(x_{1:d-1}, C)z_{d}.
\end{equation}
\vspace{-7pt}

Other flavor of normalizing flow methods builds upon invertible convolutions such as $1\times1$ in (Glow) \cite{ref:2018_Glow} and $d\times d$ in \cite{ref:2019_dxdGlow}. The work in \cite{ref:2019_cGlow} modified Glow to work with external conditions for structured output learning, yielding a non-autoregressive normalizing flow model.

\vspace{-3pt}
\section{HCNAF}
\label{sec:hcnaf}

We propose Hyper-Conditioned Neural Autoregressive Flow (HCNAF), a novel autoregressive flow where a transformation between $X = [x_{1}, x_{2}, ..., x_{D}] \in \mathbb{R}^{D}$ and $Z = [z_{1}, z_{2}, ..., z_{D}] \in \mathbb{R}^{D}$ is modeled using a non-linear neural network $f(X; \theta) = Z$ whose parameters $\theta$ are determined by arbitrarily complex conditions $C \in \mathbb{R}^{D_{c}}$ in non-autoregressive fashion, via a separate neural network $f_{H}(C) = \theta$. $f_{H}(C)$ is designed to compute the parameters for $f()$, thus being classified as an hyper-network \cite{ref:2016_HyperNetwork}. HCNAF models a conditional joint distribution $p(x_{1}, x_{2}, ... , x_{D}|C)$ autoregressively on $x_{1:D}$, by factorizing it over $D$ conditional distributions $\prod^{D}_{d=1}p(x_{d}|x_{1:d-1}, C)$. 

\textit{NAF}\cite{ref:2018_NAF} and HCNAF both use neural networks but those are different in probability modeling, conditioner network structure, and flow transformation as specified below:

\vspace{-5pt}
\begin{equation}
\label{Eq:prob_NAF}
 \left.\begin{aligned}
        p(x_{1}, x_{2}, ... , x_{D}) = \prod^{D}_{d=1}p(x_{d}|x_{1:d-1}),\\
        f_{c}(x_{1:d-1}) = \theta_{d},\\
        f(x_{d};\theta_{d}) = z_{d},
       \end{aligned}
 \right\}
 \quad \text{NAF}
\end{equation}
\vspace{-5pt}
\begin{equation}
\label{Eq:prob_HCNAF}
 \left.\begin{aligned}
        p(x_{1}, x_{2}, ... , x_{D}|C) =
        \prod^{D}_{d=1}p(x_{d}|x_{1:d-1}, C),\\
        f_{H}(C) = \theta, \theta_{d} \in \theta,\\
        f(x_{d};x_{1:d-1}, \theta_{d}) = z_{d}.
       \end{aligned}
 \right\}
 \quad \text{HCNAF}
\end{equation}

\vspace{5pt}
In Equations \ref{Eq:prob_NAF}, \textit{NAF} uses a conditioner network $f_{c}$ to obtain the parameters $\theta_{d}$ for the transformation between $x_{d}$ and $z_{d}$, which is parameterized by autoregressive conditions $x_{1:d-1}$. In contrast, in Equations \ref{Eq:prob_HCNAF}, HCNAF models the transformation to be parameterized on both  $x_{1:d-1}$, and an arbitrarily large external conditions $C$ in non-autoregressive fashion via the hyper-network $f_{H}$. For probability modeling, the difference between the two is analogous to the difference between VAE\cite{ref:2013_VAE} and conditional VAE\cite{ref:2015_cVAE}, and that between GAN\cite{ref:2014_GAN} and conditional GAN\cite{ref:2014_cGAN}. 

As illustrated in Figure \ref{fig:hcnaf_overview}, HCNAF consists of two modules: 1) a neural-net based conditional autoregressive flow, and 2) a hyper-network which computes the parameters of 1). The modules are detailed in the following sub-sections.

\begin{figure}[ht] 
    \centering
    \includegraphics[width=\linewidth]{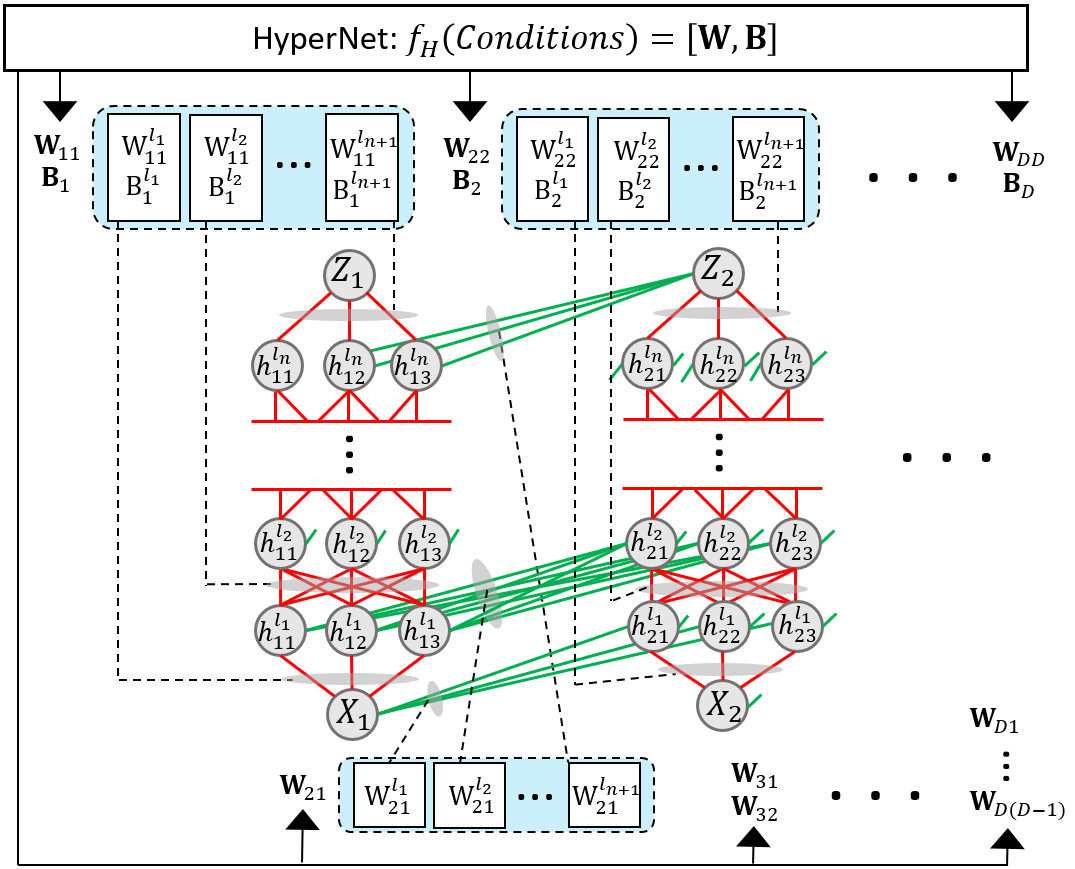}
    \caption{HCNAF's conditional AF model $f$ is a neural-net whose parameters are determined by a hyper-network $f_{H}$. The figure describes a $D$ dimensional conditional AF with $n$ hidden layers with 3 nodes. The dash lines refer to connections from $f_{H}$ to \emph{parameters} of $f$. Red lines between adjacent hidden layers $h^{l_{k-1}}_{d}$, $h^{l_{k}}_{d}$ $(\forall d, 1 \leq k \leq n+1)$ indicate that $W^{l_{k}}_{dd}$ is strictly positive. Green lines between layers $h^{l_{k-1}}_{a}$, $h^{l_{k}}_{b}$ in different flow dimensions $(1 \leq a < b \leq D, 1 \leq k \leq n+1)$ have no such constraint (i.e., $W^{l_{k}}_{ba}$ is unconstrained.)}
    \label{Fig:HCNAF_base}
\end{figure}

\subsection{NN-based Conditional Autoregressive Flow}

The proposed conditional AF is a bijective neural-network $f(X;\theta) = Z$, which models transformation between random variables $X$ and latent variables $Z$. The network parameters $\theta := [\textbf{W}, \textbf{B}]$ are determined by the hyper-network $f_{H}(C) = \theta$. The main difference between regular neural nets and flow models is the invertibility of $f^{-1}(Z) = X$ as regular networks are not typically invertible.

The conditional AF is shown in Figure \ref{Fig:HCNAF_base}. In each dimension $d$ of the flow, the bijective transformation between $x_{d}$ and $z_{d}$ are modeled with a multi-layer perceptron (MLP) with $n$ hidden layers as follows:

\vspace{-5pt}
\begin{equation}
\label{Eq:HCNAF_XtoZ}
\begin{aligned}
x_{d} \leftrightarrow h^{l_{1}}_{d} \leftrightarrow h^{l_{2}}_{d} \leftrightarrow ... \leftrightarrow h^{l_{n}}_{d} \leftrightarrow z_{d} (=h^{l_{n+1}}_{d}).
\end{aligned}
\end{equation}
\vspace{-3pt}

The connection between two adjacent hidden layers $h^{l_{k}}_{d}$ and $h^{l_{k-1}}_{d}$ is defined as:

\vspace{-5pt}
\begin{equation}
\label{Eq:HCNAF_layers}
\begin{aligned}
h^{l_{k}}_{d} 
= \phi (W^{l_{k}}_{dd}h^{l_{k-1}}_{d} +
\sum_{r=1}^{d-1}(W^{l_{k}}_{dr}h^{l_{k-1}}_{r}) +
B^{l_{k}}_{d}),
\end{aligned}
\end{equation}
\vspace{-5pt}

where subscript and superscript each denotes flow number and layer number. Specifically, $h^{l_{k}}_{d}$ is the hidden layer $l_{k}$ of the $d$-th flow. $W^{l_{k}}_{dr}$ and $B^{l_{k}}_{d}$ denote the weight matrix which defines contributions to the hidden layer $l_{k}$ of the $d$-th flow from the hidden layer $l_{k-1}$ of the $r$-th flow, and the bias matrix which defines the contributions to the hidden layer $l_{k}$ of the $r$-th flow. Finally, $\phi()$ is an activation function.

The connection between $x_{d}$ and the first hidden layer, and between the last hidden layer and $z_{d}$ are defined as:

\vspace{-3pt}
\begin{equation}
\label{Eq:HCNAF_layers_first_n_last}
\begin{aligned}
h^{l_{1}}_{d} 
= \phi (W^{l_{1}}_{dd}x_{d} +
\sum_{r=1}^{d-1}(W^{l_{1}}_{dr}x_{r}) +
B^{l_{1}}_{d}), \\[-1pt]
z_{d} 
= W^{l_{n+1}}_{dd}h^{l_{n}}_{d} + 
\sum_{r=1}^{d-1}(W^{l_{n+1}}_{dr}h^{l_{n}}_{r}) + 
B^{l_{n+1}}_{d}. \\[-1pt]
\end{aligned}
\end{equation}
\vspace{-3pt}

$h^{l_{k}}$ are the hidden units at the hidden layer $l_{k}$ across all flow dimensions $d=1:D$ and are expressed as: 

\vspace{-3pt}
\begin{equation}
\label{Eq:HCNAF_onestep_layer}
h^{l_{k}} = \phi( W^{l_{k}}h^{l_{k-1}} + B^{l_{k}} ),
\end{equation}
\vspace{-3pt}

where $W^{l_{k}}$ and $B^{l_{k}}$ are the weights and biases matrices at the hidden layer $l_{k}$ across all flow dimensions:

\vspace{-5pt}
\begin{gather}
W^{l_{k}} = 
\begin{bmatrix}
\label{Eq:HCNAF_base_WB}
    W^{l_{k}}_{11} & \textbf{0}     & \dots  & \textbf{0} \\
    W^{l_{k}}_{21} & W^{l_{k}}_{22} & \dots  & \textbf{0} \\
    \vdots         & \vdots         & \ddots & \vdots     \\
    W^{l_{k}}_{D1} & W^{l_{k}}_{D2} & \dots  & W^{l_{k}}_{DD}
\end{bmatrix},
\hspace{0.5cm} %
B^{l_{k}} = 
\begin{bmatrix}
    B^{l_{k}}_{1}  \\
    B^{l_{k}}_{2}  \\
    \vdots \\
    B^{l_{k}}_{D}
\end{bmatrix}.
\end{gather}
\vspace{-3pt}

Likewise, $\textbf{W}$ and $\textbf{B}$ denote the weights and biases matrices for all flow dimensions across all the layers. Specifically, $\textbf{W}:=\{\forall k: W^{l_{k}}\}$ and $\textbf{B}:=\{\forall k: B^{l_{k}}\}$.

Finally, $Z=f(X)$ is obtained by computing the terms from Equation \ref{Eq:HCNAF_onestep_layer} for all the network layers, from the first $X=h^{l_{0}}$ to the last layer, $Z=h^{l_{n+1}}$.

We designed HCNAF so that the hidden layer units $h^{l_{k}}_{1:D}$ are connected to the hidden units of previous layers $h^{l_{k-1}}_{1:D}$, inspired by \textit{BNAF}, as opposed to taking $h^{l_{0:n+1}}_{d}$ as inputs to a separate hyper-network to produce $h^{l_{0:n+1}}_{d+1}$ over $d=1:D$, such as presented in \textit{NAF}. This approach avoids running the hyper-network $D$ times; an expensive operation for large hyper-networks. By designing the hyper-network to output $h^{l_{0:n+1}}_{1:D}$ all at once, we reduce the computation load, while allowing the hidden states across all layers and all dimensions to contribute to the flow transformation, as $x_{d}$ is conditioned not only on $x_{1:d-1}$, but also on all the hidden layers $h^{l_{0:n+1}}_{1:d-1}$.

All Flow models must satisfy the following two properties: 1) monotonicity of $f(X)=Z$ to ensure its invertibility, and 2) tractable computation of the jacobian matrix determinant $\abs{det\frac{dZ}{dX}}$. 

\subsubsection{Invertibility of the Autoregressive Flow}
\label{Sec:Invertibility_Flow}

The monotonicity requirement is equivalent to having $\forall d: \frac{dz_{d}}{dx_{d}} > 0$, which is further factorized as:

\vspace{-2pt}
\begin{equation}
\label{Eq:HCNAF_monotonicity}
\frac{dz_{d}}{dx_{d}} 
= 
\frac{dz_{d}}{dh^{l_{n}}_{d}} \prod^{n-1}_{k=1} \frac{dh^{l_{k+1}}_{d}}{dh^{l_{k}}_{d}}
\frac{dh^{l_{1}}_{d}}{dx_{d}}
=
W^{l_{n+1}}_{dd}\prod^{n-1}_{k=0} \frac{dh^{l_{k+1}}_{d}}{dh^{l_{k}}_{d}},
\end{equation}

where $\frac{dh^{l_{k+1}}_{d}}{dh^{l_{k}}_{d}}$ $\forall k \in \{0,...,n-1\}$ is expressed as:

\vspace{-2pt}
\begin{equation}
\label{Eq:HCNAF_monotonicity2}
\frac{dh^{l_{k+1}}_{d}}{dh^{l_{k}}_{d}} 
= 
\frac{d\phi(A^{l_{k+1}}_{d})}{dA^{l_{k+1}}_{d}}\frac{dA^{l_{k+1}}_{d}}{dh^{l_{k}}_{d}}
=
\frac{d\phi(A^{l_{k+1}}_{d})}{dA^{l_{k+1}}_{d}}W^{l_{k+1}}_{dd}.
\end{equation}
\vspace{-2pt}

$A^{l_{k}}_{d}$ denotes the pre-activation of $h^{l_{k}}_{d}$. The invertibility is satisfied by choosing a strictly increasing activation function $\phi$ (e.g. \textit{tanh} or \textit{sigmoid}) and a strictly positive $W^{l_{k}}_{dd}$. $W^{l_{k}}_{dd}$ is made strictly positive by applying an element-wise exponential to all entries in $\forall d,k: W^{l_{k}}_{dd}$ at the end of the hypernetwork, inspired by \cite{ref:2019_BNAF}. Note that the operation is omitted for the non-diagonal elements of $W^{l_{k}}_{ij}, i\neq j$.

\subsubsection{Tractable Computation of Jacobian Determinant}
\label{Sec:Compute_Jacdet}

The second requirement for flow models is to efficiently compute the jacobian matrix determinant $\abs{det\frac{dZ}{dX}}$, where:

\vspace{-2pt}
\begin{equation}
\label{Eq:HCNAF_detjac}
\frac{dZ}{dX}
= \frac{dZ}{dh^{l_{n}}} \prod^{n-1}_{k=0} \frac{dh^{l_{k+1}}}{dh^{l_{k}}}
= W^{l_{n+1}} \prod^{n-1}_{k=0} \frac{d\phi(A^{l_{k+1}})}{dA^{l_{k+1}}}W^{l_{k+1}}
.
\end{equation}
\vspace{-3pt}

Since we designed $W^{l_{k+1}}$ to be lower-triangular, the product of lower-triangular matrices, $\frac{dZ}{dX}$, is also lower-triangular, whose log determinant is then simply the product of the diagonal entries: $log\abs{det\frac{dZ}{dX}}=log\abs{\prod^{D}_{d=1}\frac{dz_{d}}{dx_{d}}}=\sum^{D}_{d=1}log(\frac{dz_{d}}{dx_{d}})$, as our formulation states $\forall d: \frac{dz_{d}}{dx_{d}} > 0$. Finally, $log(\frac{dz_{d}}{dx_{d}})$ is expressed via Equations \ref{Eq:HCNAF_monotonicity} and \ref{Eq:HCNAF_monotonicity2}.

\vspace{-5pt}
\begin{equation}
\label{Eq:HCNAF_detjac_final}
log \left( \frac{dz_{d}}{dx_{d}} \right)
= log \left( W^{l_{n+1}}_{dd} \prod^{n-1}_{k=0} \frac{d\phi(A^{l_{k+1}}_{d})}{dA^{l_{k+1}}_{d}}W^{l_{k+1}}_{dd} \right)
.
\end{equation}
\vspace{-5pt}

Equation \ref{Eq:HCNAF_detjac_final} involves the multiplication of matrices in different sizes; thus cannot be broken down to a regular log summation. To resolve this issue, we utilize log-sum-exp operation on logs of the matrices in Equation \ref{Eq:HCNAF_detjac_final} as it is commonly utilized in the flow community (e.g. \textit{NAF}\cite{ref:2018_NAF} and \textit{BNAF}\cite{ref:2019_BNAF}) for numerical stability and efficiency of the computation. This approach to computing the jacobian determinant is similar to the one presented in \textit{BNAF}, as our conditional AF resembles its flow model.

As HCNAF is a member of the monotonic neural-net based autoregressive flow family like \textit{NAF} and \textit{BNAF}, we rely on the proofs presented \textit{NAF} and \textit{BNAF} to claim that HCNAF is also a universal distribution approximator. 

\subsection{Hyper-conditioning and Training}
\label{Sec:Method_HyperNet}

The key point from Equation \ref{Eq:HCNAF_XtoZ} - \ref{Eq:HCNAF_detjac_final} and Figure \ref{Fig:HCNAF_base} is that HCNAF is constraint-free when it comes to the design of the hyper-network. The flow requirements from Sections \ref{Sec:Invertibility_Flow} and \ref{Sec:Compute_Jacdet} do not apply to the hyper-network. This enables the hyper-network to grow arbitrarily large and thus to scale up with respect to the size of conditions. The hyper-network $f_{H}(C)$ can therefore be an arbitrarily complex neural network with respect to the conditions $C$.
\\

We seek to learn the target distribution $p(X|C)$ using HCNAF by minimizing the negative log-likelihood (NLL) of $p_{model}(X|C)$, i.e. the cross entropy between the two distributions, as in: 
\begin{equation}
\label{Eq:HCNAF_loss}
L := -E_{X\sim p(X|C)}[logp_{model}(X|C)] = H(p, p_{model}).
\end{equation}

Note that minimizing the NLL is equivalent to minimizing the (forward) KL divergence between the data and the model distributions $D_{KL}(p(X|C)||p_{model}(X|C))$, as $H(p,q) = H(p) + D_{KL}(p||q)$ where $H(p)$ is bounded. \\

\section{Probabilistic Occupancy Map Forecasting}
\label{sec:POM_forecasting}

In Section \ref{sec:hcnaf}, we showed that HCNAF can accommodate high-dimensional condition inputs for conditional probability density estimation problems. We leverage this capability to tackle the probabilistic occupancy map (POM) of actors in self-driving tasks. This problem operates on over one million dimensions, as spatio-temporal multi-actor images are part of the conditions. This section describes the design of HCNAF to support POM forecasting. We formulate the problem as follows:
\vspace{-3pt}
\begin{equation}
\label{eq:gen_prob_and_conditions}
p(X^{A_{i}}_{t}|C)
   \quad\mathrm{with}\quad 
C := \{ X^{A_{i}}_{-\tau:0}, X^{A_{\forall j \neq i}}_{-\tau:0}, \Omega \},
\end{equation}

where $X^{A_{i}}_{-\tau:0} \in \mathbb{R}^{\tau \times d_{i}}$ is the past states, with $d_{i}$ as the dimension of the observed state, over a time span $\tau$. $X^{A_{\forall j \neq i}}_{-\tau:0} \in \mathbb{R}^{\tau \times N_{A}}$ denotes the past states for all $N_{A}$ neighboring actors over the same time span. $\Omega \in \mathbb{R}^{N_{C} \times H \times W}$ encodes contextual static and dynamic scene information extracted from map priors (e.g. lanes and stop signs) and/or perception modules (e.g. bounding boxes for actors) onto a rasterized image of size $H$ by $W$ with $N_{C}$ channels. However comprehensive, the list of conditions in $C$ is not meant to be limitative; as additional cues are introduced to better define actors or enhance context, those are appended to the conditions. We denote $X^{A_{i}}_{t} := [x^{A_{i}}_{t}, y^{A_{i}}_{t}]$ as the location of an actor $A_{i}$ over the 2D bird's-eye view (bev) map at time $t$, by adapting our conditional AF to operate on 2 dimensions. As a result, the joint probability is obtained via autoregressive factorization given by $p(x_{t}, y_{t}|C) = p(y_{t}|x_{t}, C)p(x_{t}|C)$.

It's possible to compute $p(x_{t_{1:T}}, y_{t_{1:T}}|C)$, a joint probability over multiple time steps via Equation \ref{Eq:prob_HCNAF}, but we instead chose to compute $p(x_{t}, y_{t}|C)$ (i.e. a marginal probability distribution over a single time step) for the following reasons:

\vspace{-1pt}
\begin{enumerate}
    \item Computing $p(x_{t_{1:T}}, y_{t_{1:T}}|C)$ implies the computation of $p(x_{t},y_{t}|x_{1:t-1},y_{1:t-1},C)$ autoregressively. While this formulation reasons about the temporal dependencies between the history and the future, it is forced to make predictions on $x_{t}, y_{t}$ dependent on unobserved variables $x_{1:t-1}$ and $y_{1:t-1}$. The uncertainties of the unobserved variables have the potential to push the forecast $x_{t}, y_{t}$ in the wrong direction.
    \item The computation of $p(x_{t_{1:T}}, y_{t_{1:T}}|C)$ is intractable in nature since it requires a marginalization over all variables $t=[0,t-1]$. We note that $p(x_{t},y_{t}|C)=\int_{-\infty}^{\infty} ... \int_{-\infty}^{\infty} p(x_{1},y_{1}, ..., x_{t},y_{t}) dx_{1} ... dy_{t-1}$ is practically impossible to integrate over.
\end{enumerate}

\vspace{-1pt}
In order to predictions predictions all time $t=0:T$, we simply incorporate a time variable as part of the conditions.

In addition to POMs, HCNAF can be used to sample trajectories using the inverse transformation $f^{-1}: Z \thicksim N(\textbf{0}_{D}, \textbf{I}_{DxD}) \rightarrow X$.
The exact probabilities of the generated trajectories can be computed via Equation \ref{Eq:Change_of_Variables}. However, it is not trivial to obtain the inverse flow since a closed form solution is not available. A solution is to use a numerical approximation or to modify the conditional AF of HCNAF; which is not discussed in this work.

\section{Experiments}
\label{sec:experiment}

In this paper, five experiments (including three experiments on publicly available datasets) of various tasks and complexities are presented to evaluate HCNAF\protect\footnotemark. For all, we provide quantitative (NLL, $D_{KL}$) and qualitative measures (visualizations; except MNIST as the dimension is large). We start the section by demonstrating the effectiveness of HCNAF on density estimation tasks for two \textit{Toy Gaussians}. We then verify the scalability of HCNAF by tackling more challenging, high dimensional ($C > \mathbb{R}^{1,000,000}$) POM forecasting problems for autonomous driving. For POM forecasting, we rely on two datasets: 1) \textit{Virtual Simulator}: a simulated driving dataset with diverse road geometries, including multiple road actors designed to mimic human drivers. The scenarios are based on real driving logs collected over North-American cities. 2) \textit{PRECOG-Carla}: a publicly available dataset created using the open-source Carla simulator for autonomous driving research \cite{ref:2019_PRECOG}. Lastly, we run a conditional density estimation task on MNIST which is detailed in the supplementary material.

\subsection{Toy Gaussians}
\label{sec:toy_gaussians}

We conducted two experiments to demonstrate the performance of HCNAF for density estimations. The first is an experiment from \textit{NAF} paper \cite{ref:2018_NAF}, and aims to show the model's learning ability for three distinct probability distributions over a 2D grid map, $p(x, y)$. The non-linear distributions are spatially distinct groups of gaussians. In the second experiment, we demonstrate how HCNAF can generalize its outputs over previously \emph{unseen} conditions.
\vspace{-4pt}

\subsubsection{Toy Gaussians: Experiment 1}
\vspace{-4pt}
\begin{figure}[ht] 
    \centering
    \includegraphics[width=0.95\linewidth]{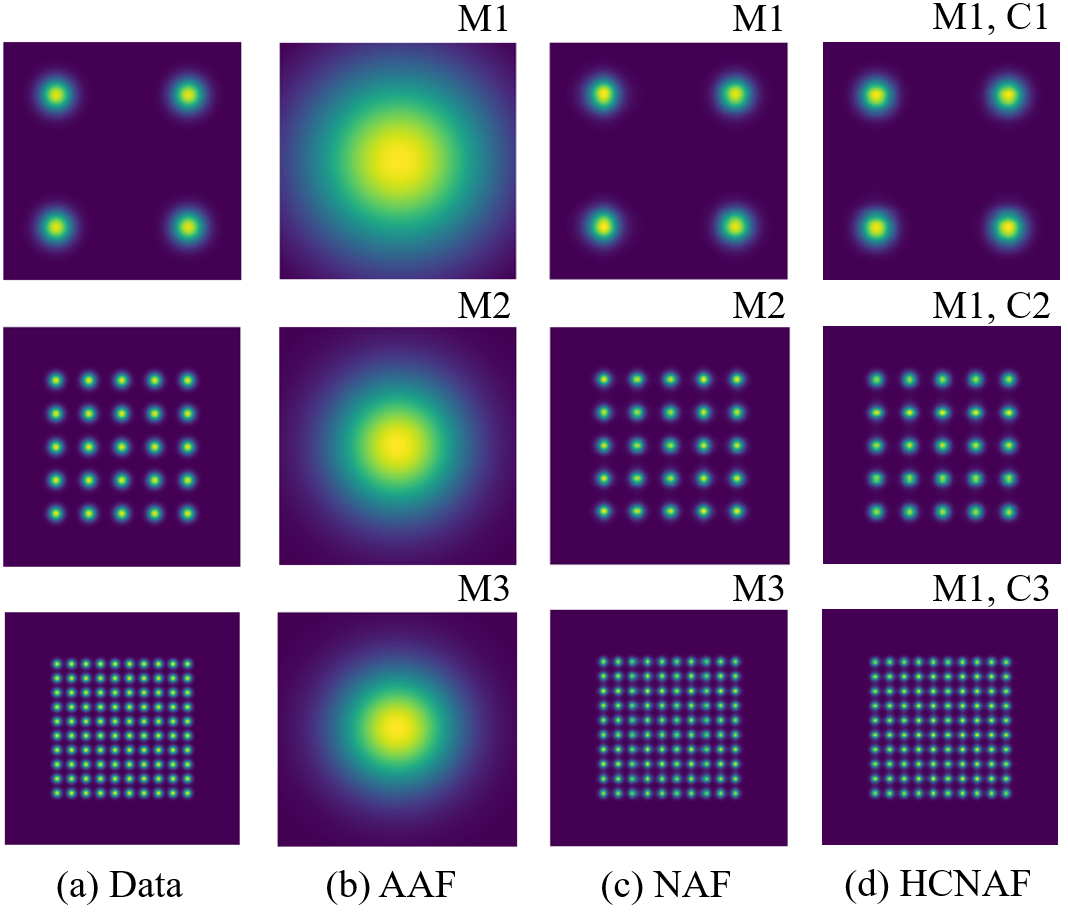}
    \caption{Density estimation tasks using three gaussian distributions. In order to reproduce the probability distributions $p(x,y)$, HCNAF uses a \textbf{single} model and three conditions, whereas NAF requires three \textbf{different} models, i.e. trained separately. In the figure, M: model and C: condition.}
    \label{Fig:NAF_Exp_Gaussians}
\end{figure}

\begin{table}[ht]
\caption{NLL for the experiment depicted in Figure \ref{Fig:NAF_Exp_Gaussians}. Lower values are better.}
\label{Table:Exp_NAF}
\begin{tabular}{
>{\centering\arraybackslash}m{1.5cm}
>{\centering\arraybackslash}m{1.4cm} 
>{\centering\arraybackslash}m{1.4cm}
>{\centering\arraybackslash}m{2.3cm} }
\toprule
 & AAF & \textbf{NAF} & HCNAF (ours) \\
\toprule
2 by 2   & 6.056 & \textbf{3.775} & 3.896 \\
\midrule
5 by 5   & 5.289 & \textbf{3.865} & 3.966 \\
\midrule
10 by 10 & 5.087 & \textbf{4.176} & 4.278 \\
\bottomrule
\end{tabular}
\end{table}

\footnotetext{The code is available at \url{https://github.com/gsoh/HCNAF}}
Results from Figure \ref{Fig:NAF_Exp_Gaussians} and Table \ref{Table:Exp_NAF} show that HCNAF is able to reproduce the three nonlinear target distributions, and to achieve comparable results as those using \textit{NAF}, albeit with a small increase in NLL. We emphasise that HCNAF uses a \textbf{single} model (with a 1-dimensional condition variable) to produce the 3 distinct pdfs, whereas AAF (Affine AF) and \textit{NAF} used 3 distinctly trained models. The autoregressive conditioning applied in HCNAF is the same as for the other two models. The hyper-network of HCNAF uses $C \in \{0,1,2\} $ where each value represents a class of 2-by-2, 5-by-5, and 10-by-10 gaussians. 

\subsubsection{Toy Gaussians: Experiment 2}

\begin{figure}[ht] 
    \centering
    \includegraphics[width=\linewidth]{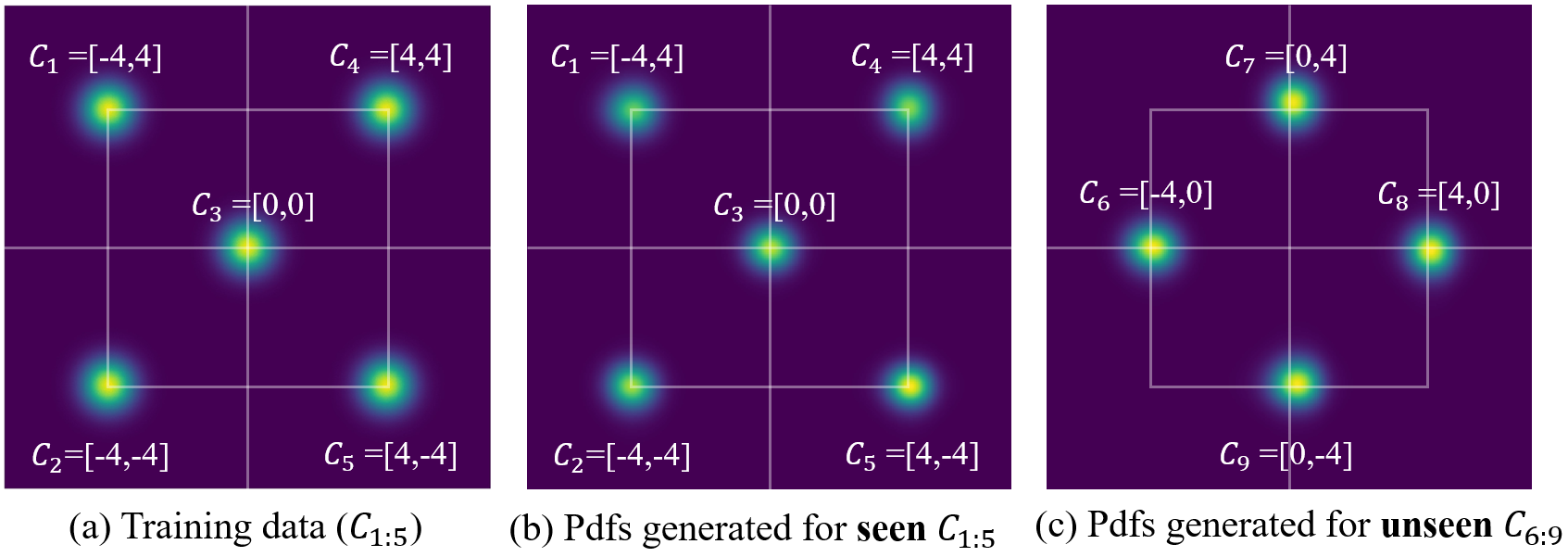}
    \caption{HCNAF model trained with 5 different discrete conditions $C_{train} = \{C_{1}, ... , C_{5}\}$, where $C_{i}$ represents the mean of an isotropic bivariate gaussian pdf. a) $p(x,y|C_{train})$,  b) $p_{model}(x,y|C_{train})$ c) predictions on previously \textbf{unseen} conditions $p_{model}(x,y|C_{unseen})$, $C_{unseen}:=\{C_{6}, ... , C_{9}\}$.}
    \label{Fig:HCNAF_Gaussians_Generalization}
\end{figure}

From the density estimation experiment shown in Figure \ref{Fig:HCNAF_Gaussians_Generalization}, we observed that HCNAF is capable of \textit{generalization} over \textbf{unseen} conditions, i.e. values in the condition terms that were intentionally omitted during training. The experiment was designed to verify that the model would interpolate and/or extrapolate probability distributions beyond the set of conditions it was trained with, and to show how effective HCNAF is at reproducing both the target distribution $p(x,y|C_{i})$ for $C_{i} \in C_{train}$. As before, we trained a single HCNAF model to learn 5 distinct pdfs, where each pdf represents a gaussian distribution with its mean (center of the 2D gaussian) used as conditions $C:=(x_{c}, y_{c}) \in \mathbb{R}^{2}$ and with an isotropic standard deviation $\sigma$ of 0.5. 

\begin{table}[ht]
\caption{Differences between the target and predicted distributions in terms of cross entropy and KL divergence for Figure \ref{Fig:HCNAF_Gaussians_Generalization}.}
\label{Table:HCNAF_Gaussians_Generalization}
\begin{tabular}{
>{\centering\arraybackslash}p{1.7cm}
>{\centering\arraybackslash}p{1.3cm} 
>{\centering\arraybackslash}p{1.8cm}
>{\centering\arraybackslash}p{1.8cm} }
\toprule
& \bm{$p(x, y)$} & \multicolumn{2}{c}{\bm{$p_{HCNAF}(x,y|C_{i})$}} \\
\toprule
$C$ & - &  $C_{i} \in C_{train}$ & $C_{i} \in C_{unseen}$ \\
\midrule
$H(p)$  & 1.452 & - & - \\
\midrule
$H(p,p_{model})$  & - & 1.489 & 1.552 \\
\midrule
$D_{KL}(p||p_{model})$  & - & 0.037 & 0.100 \\
\bottomrule
\end{tabular}
\end{table}

For this task, the objective function is the maximization of log-likelihood, which is equivalent to the maximization of the KL divergence $-E_{(x,y)\sim N(C_{i},0.25\cdot I)}[logp_{model}(x,y|C_{i})]$ where $C_{i}$ is uniformly sampled from the set of conditions $C_{train} := \{C_{1}, C_{2}, ... , C_{5}\}$. Table \ref{Table:HCNAF_Gaussians_Generalization} provides quantitative results from the cross entropy $H(p, p_{model})$ and a KL divergence $D_{KL}(p||p_{model})$. Note that $H(p, p_{model})$ is lower-bounded by $H(p)$ since $H(p, p_{model}) = H(p) + D_{KL}(p||p_{model})$. The differential entropy $H(p)$ of an isotropic bi-variate Gaussian distribution $p(x,y)$ and is computed using: $H(p)=0.5\cdot ln(2\pi e(\sigma)^2)^2$. The results show that HCNAF is able to generalize its predictions for unseen conditions as shown by the small deviation of $H(p,p_{model})$ from its lower bound $H(p)$.
\\

\subsection{Forecasting POM for Autonomous Driving}

Through changes in the hyper-network, we show how HCNAF can be scaled up to tackle the POM forecasting problems for autonomous driving. The condition $C$ is now significantly larger when compared to that from the experiments in Section \ref{sec:toy_gaussians}, as shown in Equation \ref{eq:gen_prob_and_conditions}. $C$ now includes information extracted from various sensors (lidar, camera), maps (lanes, stop-signs), and perception object detections (expressed as bounding boxes for actors), with a total dimension is in the millions of parameters. As per its design, HCNAF's AF network is unaffected by the increase in conditional dimensions.

\begin{figure}[ht] 
    \centering
    \includegraphics[width=\linewidth]{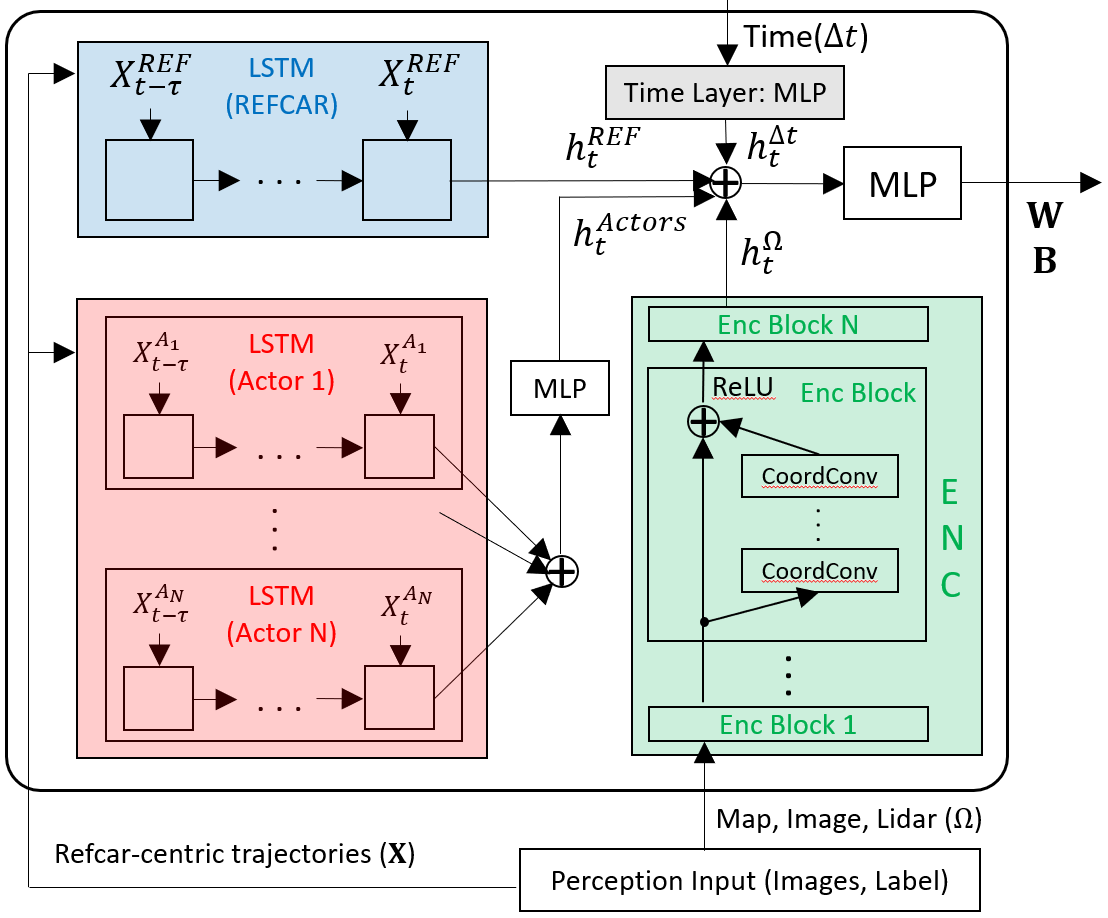}
    \caption{Design for the Hyper-network of HCNAF used in the POM forecasting problem.}
    \label{Fig:HCNAF_POM_model}
\end{figure}

Figure \ref{Fig:HCNAF_POM_model} depicts the customized hyper-network used for the POM forecasting task. The hyper-network takes perception inputs as the condition $C$, and outputs a set of network parameters \textbf{W} and \textbf{B} for the subsequent HCNAF's conditional AF $f(\cdot ; \textbf{W}, \textbf{B}): X \leftrightarrow Z \sim N(0,I_{2x2})$. The inputs come from various sensors (lidar or camera) through a perception module and also from prior map information. Specifically, $C$ is formed with 1) the bev images which include lanes, stop-signs, lidar data, and actor bounding boxes in a 2D grid map (see figures presented in the supplementary material and Figure \ref{Fig:PRECOG_POM_Main}), and 2) the states of actors in actor-centric pixel coordinates. The perception module used reflects other standard approaches for processing multi-sensor data, such as \cite{ref:2015_ResNet}. The hyper-network consists of three main components: 1) LSTM modules, 2) an \textit{encoder} module, and 3) a \textit{time} module. The outputs of the three modules $h^{REF}_{t}, h^{Actors}_{t}, h^{\Omega}_{t}, h^{\Delta t}_{t}$ are concatenated and fed into an MLP, which outputs \textbf{W} and \textbf{B}, as shown in Figure \ref{Fig:HCNAF_POM_model}. 

The \textit{LSTM module} takes the states of an actor $A_{i}$ in the scene $X^{A_{i}}_{t-\tau: t}$ where $X^{A_{i}}_{t} := [x^{A_{i}}_{t}, y^{A_{i}}_{t}, sin(\theta^{A_{i}}_{t}), cos(\theta^{A_{i}}_{t}), v^{A_{i}}]$ to encode temporal dependencies and trends among the state parameters. A total of $N+1$ LSTM modules are used to model the $N$ actors and the reference car for which we produce the POM forecasts. The resulting outputs are $h^{REF}_{t} \in \mathbb{R}^{d_{REF}=20}$, and $h^{Actors}_{t} \in \mathbb{R}^{d_{Actors}=18}$.

The \textit{encoder module} takes in the bev images denoted as $\Omega$. The role of this module is to transform the scene contexts into a one-dimensional tensor that is concatenated with other parameters of our conditional AF flow module. We use residual connections to enhance the performance of our encoder as in \cite{ref:2015_ResNet}. Since our hyper-network works with Cartesian (x,y) space and pixel (image) space, we use coordinate convolution (coordconv) layers as in \cite{ref:2018_CoordConv} to strengthen the association between the two data. Overall, the \textit{encoder} network consists of 4 encoder blocks, and each encoder block consists of 5 coordconv layers with residual connections, max-pooling layers, and batch-normalization layers. The resulting output is $h^{\Omega}_{t} \in \mathbb{R}^{d_{\Omega}=64}$.

Lastly, the \textit{time layer} adds the forecasting time $\Delta t \in \mathbb{R}^{1}$, i.e. time span of the future $t$ away from the reference (or present) time $t=0$. In order to increase the contribution of the time condition, we apply an MLP which outputs a hidden variable for the time condition $h^{\Delta t}_{t} \in \mathbb{R}^{d_{t}=10}$. 

\subsubsection*{Forecasting POM with a Virtual Simulator Dataset}

\begin{figure*}[ht] 
    \centering
    \includegraphics[width=0.97\linewidth]{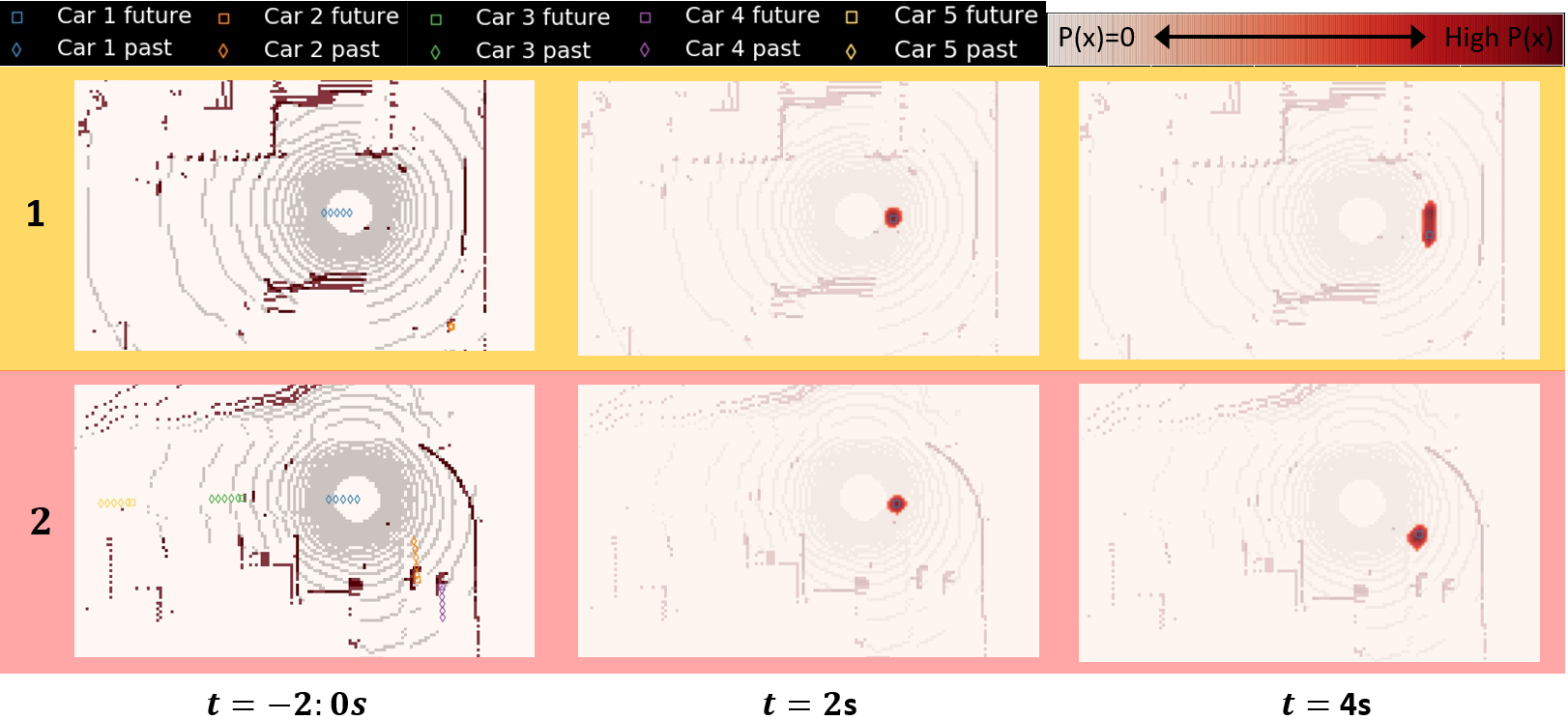}
    \caption{Visualized POM forecasts on \textit{PRECOG-Carla} dataset using the HCNAF model described in Table \ref{Table:Exp_PRECOG_Carla} (with lidar). Left: 2 seconds history of cars. Center and right: probabilistic occupancy predictions for Car 1 at $t$ = 2 and 4 secs depicted as red heatmaps, with actor ground truth (blue square) overlayed. Note that we only forecast POMs for the car 1 as the lidar data is only available for the car 1. In the examples 1, the car 1 enters a 3-way intersection, HCNAF uses the road geometry coming from the lidar data and correctly forecasts that there are two natural modes (left-turn \& right-turn) and depict the probabilities of positions as heatmaps. In example 2, HCNAF uses the curved road geometry and successfully forecasts the occupancy probabilities of the car 1. More results on POM visualizations are presented in the supplementary material.}
    \label{Fig:PRECOG_POM_Main}
\end{figure*}

Using the POM hyper-network, HCNAF was trained on an internal dataset that we call \textit{Virtual Simulator}. The dataset is comprised of bev images of size $N \times 256 \times 256$, where $N$ may include all, or a subset of the following channels: stop signs, street lanes, reference car locations, and a number of actors. We also add the history of actor states in pixel coordinates, as discussed in the previous sub-section. For each of the vehicles/actors, we apply a coordinate transformation to obtain actor-centric labels and images for training. The vehicle dataset includes parked vehicles and non-compliant road actors to introduce common and rare events (e.g. sudden lane changes or sudden stopping in the middle of the roads). We produce POM for all visible vehicles, including parked vehicles, and non-compliant actors, even if those are not labeled as such. Note that the dataset was created out of several million examples, cut into snippets of 5 seconds in duration. We present a figure which depicts POM forecasts for three scenarios sampled from the test set and a table for an ablation study to show the impact of different hyper-networks inputs on the POM forecasting accuracy in the supplementary material.

As discussed in Section \ref{sec:POM_forecasting}, HCNAF produces not only POM, but also trajectory samples via the inverse transformation of the conditional AF $f^{-1}$. As we advocate the POM approach, we do not elaborate further on the trajectory based approach using HCNAF.

\subsubsection*{Forecasting POM with PRECOG-Carla Dataset}

We trained HCNAF for POM forecasting on the PRECOG-Carla Town01-train dataset and validated the progress over Town01-val dataset \cite{ref:2019_PRECOG}. The hyper-network used for this experiment was identical to the one used for the \textit{Virtual Simulator} dataset, except that we substituted the bev images with two overhead lidar channels; the above ground and ground level inputs. The \textit{encoder} module input layer was updated to process the lidar image size (200x200) of the PRECOG-Carla dataset. In summary, $C$ included the lidar data, and the history of the reference car and other actors.

To evaluate the performance of the trained models, \cite{ref:2019_PRECOG} used the \textit{extra nats} $\hat{e}$ metric for the likelihood estimation instead of NLL. $\hat{e}$ is a normalized, bounded likelihood metric defined as $\hat{e}:=[H(p',p_{model})-H(\eta)]/(T \cdot A \cdot D) \geq 0$, where $H(p',p_{model}), T, A, D$ each represents the cross-entropy between $p'$ (perturbed with an isotropic gaussian noise) and $p_{model}$, prediction horizon, number of actors, and dimension of the actor position. We used the same $\eta = N(\textbf{0}, 0.01^{2}\cdot \textbf{I})$ as cited, whose differential entropy is analytically obtained using $H(\eta)=0.5\cdot T\cdot A\cdot D\cdot ln(2\pi e|\Sigma|)$. We computed $p(x_{t}, y_{t}|C)$ over all time-steps available in the dataset. The results are presented in Table \ref{Table:Exp_PRECOG_Carla} and Figure \ref{Fig:PRECOG_POM_Main}. 

It is worth mentioning that there exists works including \cite{ref:2019_Multiple_Pred}, \cite{ref:2019_multipath} that used the PRECOG-Carla dataset. However, most reported trajectory-based predictions metrics (MSE, MinMSD, etc). To the best of our knowledge, the only available benchmark for \emph{NLL} on the PRECOG dataset is what is presented in this paper (PRECOG-ESP). Since we take the occupancy-based approach, trajectory-based metrics are not applicable to our approach. 

\begin{table}[h]
\caption{PRECOG-CARLA Town01 Test, 1 agent, mean $\hat{e}$}
\label{Table:Exp_PRECOG_Carla}
\begin{tabular}{ 
p{3.5cm} 
p{4.0cm}
}
\toprule
\textbf{Method} & \textbf{Test ($\hat{e}$): Lower is better} \\
\toprule
PRECOG-ESP, no lidar   & 0.699 \\
\midrule
PRECOG-ESP             & 0.634 \\
\midrule
HCNAF, no lidar (ours)        & 0.184 \\
\midrule
\textbf{HCNAF} (ours)         & \textbf{0.114} (5+ times lower) \\
\bottomrule
\end{tabular}
\end{table}

We believe that HCNAF performed better than \textit{PRECOG}\textit{-ESP}, which is a state-of-the-arts prediction model in autonomous driving, by taking advantage of HCNAF's expressivity comes from non-linear flow transformations and having condition terms affecting the hidden states of all layers of HCNAF's conditional AF. Note, \textit{PRECOG} utilizes bijective transformations $f:X \leftrightarrow Z$ that is rooted in affine AF, similar to \textit{cMAF} (See Equation \ref{Eq:flow_transformation_cMAF}). We also believe that the HCNAF's generalization capability is a contributing factor that explains how HCNAF is able to estimate probability densities conditioned on previously unseen contexts.

\vspace{-8pt}
\section{Conclusion}

We present HCNAF, a novel universal distribution approximator tailored to model conditional probability density functions. HCNAF extends neural autoregressive flow \cite{ref:2018_NAF} to take arbitrarily large conditions, not limited to autoregressive conditions, via a hyper-network which determines the network parameters of HCNAF's AF. By modeling the hyper-network constraint-free, HCNAF enables it to grow arbitrarily large and thus to scale up with respect to the size of non-autoregressive conditions. We demonstrate its effectiveness and capability to generalize over unseen conditions on density estimation tasks. We also scaled HCNAF's hyper-network to handle larger conditional terms as part of a prediction problem in autonomous driving. 

{\small
\bibliographystyle{unsrt} 
\bibliography{egbib}
}

\clearpage

\beginsupplement
\section{Ablation Study on the POM forecasting experiment with Virtual Simulator Dataset}
\label{sec:Vsim_Ablation_Study}

In this section, we present the results of an ablation study conducted on the test set (10\% of the dataset) of the \textit{Virtual Simulator} experiment to investigate the impact of different hyper-networks inputs on the POM forecasting performance. As mentioned in Section \ref{sec:POM_forecasting} and \ref{sec:toy_gaussians}, each 5-second snippet is divided into a 1-second long history and a 4-second long prediction horizons. All the inputs forming the conditions $C$ used in the ablation study are extracted from the history portion, and are listed below.

\begin{enumerate}
   \item $X^{REF}_{t=-1:0}$: historical states of the reference car in pixel coordinates,
   \item $X^{A_{i}}_{t=-1:0}$: historical states of the actors excluding the reference car in pixel coordinates and up to 3 closest actors,
   \item $X^{SS}_{t=0}$: stop-sign locations in pixel coordinates, and
   \item $\Omega \in \{\emptyset, \Omega_{All}\}$: bev images of size $N \times 256 \times 256$. $N$ may include all, or a subset of the following channels: stop signs at $t=0$, street lanes at $t=0$, reference car \& actors images over a number of time-steps $t\in[-1,0]s$. 
\end{enumerate}

\begin{table*}[ht]
\caption{Ablation study on \textit{Virtual Simulator}. The evaluation metric is negative log-likelihood. Lower values are better.}
\label{Table:Ablation_study_Vsim}
\begin{tabular}{ 
p{1.3cm} 
p{1.3cm}
p{1.9cm} 
p{1.9cm}
p{1.9cm} 
p{1.9cm}
p{1.9cm} 
p{1.9cm}
}
\toprule
\multicolumn{2}{c}{\multirow{2}{*}{\textbf{Conditions}}} 
& \multicolumn{4}{c}{$\Omega=\emptyset$}
& \multicolumn{2}{c}{$\Omega=\Omega_{All}$}
\\
\cmidrule{3-8}
& & $C_{1}$ & $C_{2}$ & $C_{3}$ & $C_{4}$ & $C_{5}$ & $C_{6}$
\\
\toprule
\multirow{2}{*}{\textbf{NLL}} & \textbf{t = 2s} 
& -8.519 & -8.015 & -7.905 & -8.238 & \textbf{-8.943} & -8.507
\\
& \textbf{t = 4s} 
& -6.493 & -6.299 & -6.076 & -6.432 & \textbf{-7.075} & -6.839
\\

\midrule

&
\multicolumn{2}{l}{$C_{1} = X^{REF}_{t-\tau:t}$} &
\multicolumn{2}{l}{$C_{3} = X^{REF}_{t-\tau:t} + X^{A_{1:N}}_{t-\tau:t}$} &
\multicolumn{3}{l}{$C_{5} = X^{REF}_{t-\tau:t} + \Omega_{All}$ (best model) }
\\[2pt]

&
\multicolumn{2}{l}{$C_{2} = X^{REF}_{t-\tau:t} + X^{A_{1:N}}_{t}$} &
\multicolumn{2}{l}{$C_{4} = X^{REF}_{t-\tau:t} + X^{A_{1:N}}_{t-\tau:t} + X^{SS}_{t}$} & 
\multicolumn{3}{l}{$C_{6} = X^{REF}_{t-\tau:t} + \Omega_{All} + X^{A_{1:N}}_{t-\tau:t} + X^{SS}_{t}$}
\\[2pt]

\bottomrule
\end{tabular}
\end{table*}

As presented in Table \ref{Table:Ablation_study_Vsim}, we trained 6 distinct models to output $p(X_{t}|C)$ for $t=2s, 4s$. The six models can be grouped into two different sets depending on the $\Omega$ that was used. The first group is the models that do not utilize any bev map information, therefore $\Omega=\emptyset$. The second group leverages all bev images $\Omega=\Omega_{All}$. Each group can be divided further, depending on whether a model uses $X^{A_{i}}_{t=-1:0}$ and $X^{SS}_{t=0}$. The HCNAF model which takes all bev images $\Omega=\Omega_{All}$ from the perception model as the conditions $C_{5}$ (see Table 1) excluding the historical states of the actors and stop-signs is denoted by the term \textit{best model}, as it reported the lowest NLL. We use $M_{i}$ to represent a model that takes $C_{i}$ as the conditions.

Note that the hyper-network depicted in Figure \ref{Fig:HCNAF_POM_model} is used for the training and evaluation, but the components of the hyper-network changes depending on the conditions. We also stress that the two modules of HCNAF (the hyper-network and the conditional AF) were trained jointly. Since the hyper-network is a regular neural-network, it's parameters are updated via back-propagations on the loss function.

As shown in Table \ref{Table:Ablation_study_Vsim}, the second group ($M_{5:6}$) performs better than the first group ($M_{1:4}$). Interestingly, we observe that the model $M_{1}$ performs better than $M_{2:4}$. We suspect that this is due to $M_{2:4}$ using \textit{imperfect} perception information. That is, not all the actors in the scene were detected and some actors are only partially detected; they appeared and disappeared over the time span of 1-second long history. The presence of non-compliant, or abnormal actors may also be a contributing factor. When comparing $M_{2}$ and $M_{3}$ we see that the historical information of the surrounding actors did not improve performance. In fact, the model that only utilizes $X^{A_{i}}$ at time $t=0$ performs better than the one using $X^{A_{i}}$ across all time-steps. Finally, having the stop-sign locations as part of the conditions is helping, as many snippets covered intersection cases. When comparing $M_{5}$ and $M_{6}$, we observe that adding the states of actors and stop-signs in pixel coordinates to the conditions did not improve the performance of the network. We suspect that it is mainly due to the same reason that $M_{1}$ performs better than $M_{4}$.

\section{Implementation Details on Toy Gaussian Experiments}
\label{sec:Toy_Exp_Implementation_Detail}

For the toy gaussian experiment 1, we used the same number of hidden layers (2), hidden units per hidden layer (64), and batch size (64) across all autoregressive flow models AAF, \textit{NAF}, and HCNAF. For \textit{NAF}, we utilized the conditioner (transformer) with 1 hidden layer and 16 sigmoid units, as suggested in \cite{ref:2018_NAF}. For HCNAF, we modeled the hyper-network with two multi-layer perceptrons (MLPs) each taking a condition $C \in \mathbb{R}^1$ and outputs \textbf{W} and \textbf{B}. Each MLP consists of 1 hidden layer, a ReLU activation function. All the other parameters were set identically, including those for the Adam optimizer (the learning rate $5e^{-3}$ decays by a factor of 0.5 every 2,000 iterations with no improvement in validation samples). The NLL values in Table \ref{Table:Exp_NAF} were computed using 10,000 samples.

For the toy gaussian experiment 2, we used 3 hidden layers, 200 hidden units per hidden layer, and batch size of 4. We modeled the hyper-network the same way we modeled the hyper-network for the toy gaussian experiment 1. The NLL values in Table \ref{Table:HCNAF_Gaussians_Generalization} were computed using 10,000 test samples from the target conditional distributions.

\section{Number of Parameters in HCNAF}
\label{sec:Num_param}

In this section we discuss the computational costs of HCNAF for different model choices. We denote $D$ and $L_{F}$ as the flow dimension (the number of autoregressive inputs) and the number of hidden layers in a conditional AF. In case of $L_{F} = 1$, there exists only 1 hidden layer $h_{l_{1}}$ between $X$ and $Z$. We denote $H_{F}$ as the number of hidden units in each layer per flow dimension of the conditional AF. Note that the outputs of the hyper-network are \textbf{W} and \textbf{B}. The number of parameters for \textbf{W} of the conditional AF is $N_{W} = D^{2}H_{F}(2 + (L_{F}-1)H_{F})$ and that for \textbf{B} is $N_{B} = D(H_{F}L_{F}+1)$.

The number of parameters in HCNAF's hyper-network is largely dependent on the scale of the hyper-network's neural network and is independent of the conditional AF except for the last layer of the hyper-network as it is connected to \textbf{W} and \textbf{B}. The term $N_{1:L_{H}-1}$ represents the total number of parameters in the hyper-network up to its $L_{H}-1$th layer, where $L_{H}$ denotes the number of layers in the hyper-network. $H_{L_{H}}$ is the number of hidden units in the $L_{H}$th (the last) layer of the hyper-network. Finally, the number of parameters for the hyper-network is given by $N_{H} = N_{1:L_{H}-1} + H_{L_{H}}(N_{W} + N_{B})$.

The total number of parameters in HCNAF is therefore a summation of $N_{W}$, $N_{B}$, and $N_{H}$. The dimension grows quadratrically with the dimension of flow $D$, as well as $H_{F}$ for $L_{F} \geq 2$. The key to minimizing the number of parameters is to keep the dimension of the last layer of the hyper-network low. That way, the layers in the hyper-network, except the last layer, are decoupled from the size of the conditional AF. This allows the hyper-network to become large, as shown in the POM forecasting problem where the hyper-network takes a few million dimensional conditions.

\section{Conditional Density Estimation on MNIST}
\label{sec:EXP_MNIST}

The primary use of HCNAF is to model conditional probability distributions $p(x_{1:D}|C)$ when the dimension of $C$ (i.e., inputs to the hyper-network of HCNAF) is large. For example, the POM forecasting task operates on large-dimensional conditions with $D_{C} > $ 1 million and works with small autoregressive inputs $D$ = 2. Since the parameters of HCNAF's conditional AF grows quickly as $D$ increases (see Section \ref{sec:Num_param}), and since the conditions $C$ greatly influence the hyper-parameters of conditional AF module (Equation 4), HCNAF is ill-suited for density estimation tasks with $D >> D_{C}$. Nonetheless, we decided to run this experiment to verify that HCNAF would compare well with other recent models. Table \ref{Table:Exp_MNIST} shows that HCNAF achieves the state-of-art performance for the conditional density estimation.

MNIST is an example where the dimension of autoregressive variables ($D$ = 784) is large and much bigger than $D_{C}$ = 1. MNIST images (size 28 by 28) belong to one of the 10 numeral digit classes. While the unconditional density estimation task on MNIST has been widely studied and reported for generative models, the \textbf{conditional} density estimation task has rarely been studied. One exception is the study of conditional density estimation tasks presented in \cite{ref:2017_MAF}. In order to compare the performance of HCNAF on MNIST ($D >> D_{C}$), we followed the experiment setup from \cite{ref:2017_MAF}. It includes the dequantization of pixel values and the translation of pixel values to logit space. The objective function is to maximize the joint probability over $X:=x_{1:784}$ conditioned on classes $C_{i} \in \{0, ..., 9\}$ of $X$ as follows.

\vspace{-8pt}
\begin{equation}
\label{Eq:MNIST}
p(x_{1:784}|C_{i})
=\prod_{d=1}^{784}p(x_{d}|x_{1:d-1},C_{i}).
\end{equation}

\vspace{-9pt}
\begin{table}[ht]
\caption{Test negative log-likelihood (in nats, logit space) and bits per pixel for the conditional density estimation task on \textbf{MNIST}. Lower values are better. Results from models other than HCNAF were found in \cite{ref:2017_MAF}. HCNAF is the best model among the conditional flow models listed.}
\label{Table:Exp_MNIST}
\begin{tabular}{P{2.17cm}P{2.6cm}P{2.23cm}}
\toprule
\textbf{Models} & \textbf{Conditional NLL} & \textbf{Bits Per Pixel}   \\
\toprule
Gaussian & 1344.7 & 1.97 \\
\midrule
MADE     & 1361.9 & 2.00 \\
\midrule
MADE MoG & 1030.3 & 1.39 \\
\midrule
Real NVP (5) & 1326.3 & 1.94 \\
\midrule
Real NVP (10) & 1371.3 & 2.02 \\
\midrule
MAF (5) & 1302.9 & 1.89 \\
\midrule
MAF (10) & 1316.8 & 1.92 \\
\midrule
MAF MoG (5) & 1092.3 & 1.51 \\
\midrule
\textbf{HCNAF} (ours) & \textbf{975.9} & \textbf{1.29} \\
\bottomrule
\end{tabular}
\end{table}

\vspace{-4pt}
For the evaluation, we computed the test log-likelihood on the joint probability $p(x_{1:784})$ as suggested in \cite{ref:2017_MAF}. That is, $p(x_{1:784})$ = $\sum_{i=0}^{9}p(x_{1:784}|C_{i})p(C_{i})$ with $p(C_{i}) = 0.1$, which is a uniform prior over the 10 distinct labels. Accordingly, the bits per pixel was converted from the LL in logit space to the bits per pixel as elaborated in \cite{ref:2017_MAF}.

For the HCNAF presented in Table \ref{Table:Exp_MNIST}, we used $L_{F}=1$ and $H_{F}=38$ for the conditional AF module. For the hyper-network, we used $L_{H}=1$, $H_{H,W}=10$ for \textbf{W}, $H_{H,B}=50$ for \textbf{B}, and 1-dimensional label as the condition $C \in \mathbb{R}^{1}$.

\section{Detailed Evaluation Results and Visualization of POMs for PRECOG-Carla Dataset}
\label{sec:PRECOG_Visualization}

\begin{figure*}[ht] 
    \centering
    \includegraphics[width=0.6\linewidth]{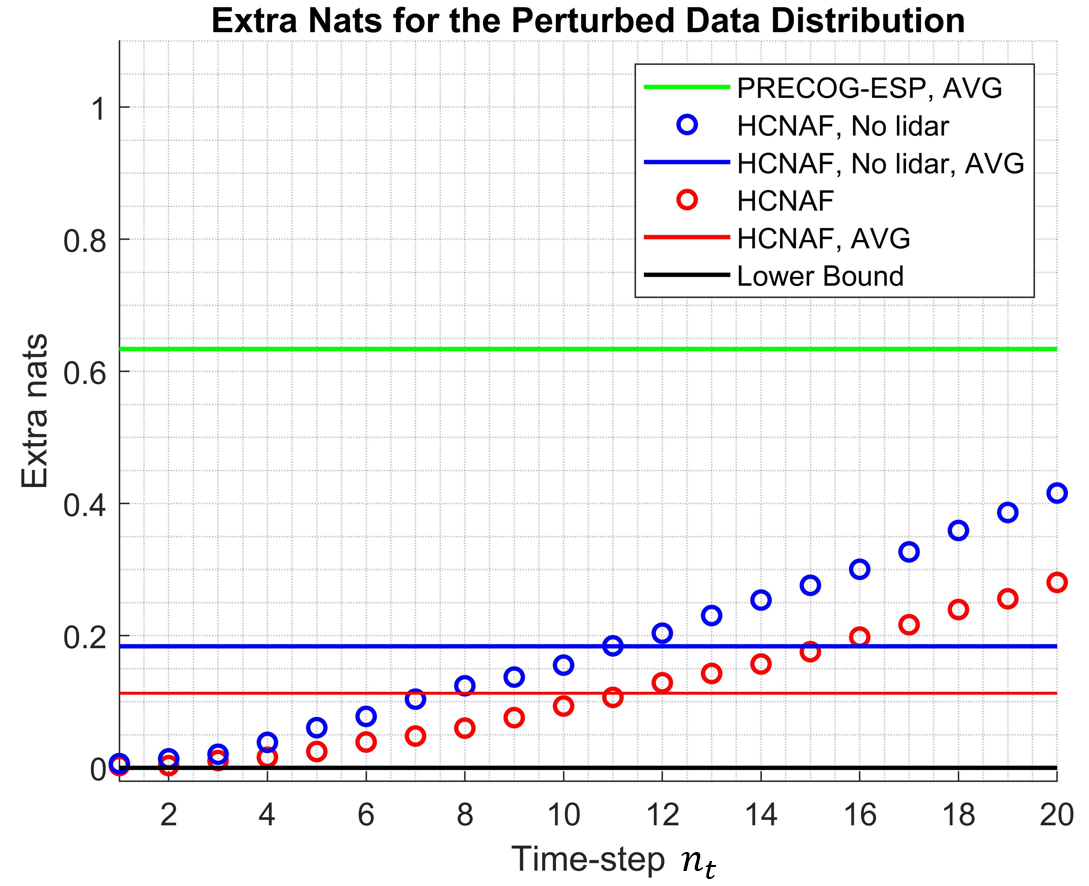}
    \caption{Detailed evaluation results on the \textit{PRECOG-Carla} test set per time-step for the \textit{HCNAF} and \textit{HCNAF(No lidar)} models, compared to average \textit{PRECOG} published performance, such as described in Table \ref{Table:Exp_PRECOG_Carla}. AVG in the plot indicates the averaged extra nats of a model over all time-steps $n_{t=1:20}$ (i.e., $\Sigma^{20}_{t=1}\hat{e}_{n_{t}}/20$). Note that the x-axis time steps are 0.2 seconds apart, thus $n_{t}=20$ corresponds to $t$ = 4 seconds into the future and that there is no upper bound of $\hat{e}$ as $\hat{e} \geq 0$. As expected, the POM forecasts $p_{model}(X|C)$ are more accurate (closer to the target distribution $p'(X|C)$) at earlier time-steps, as the uncertainties grow over time. For all time-steps, the HCNAF model with lidar approximates the target distribution better than the HCNAF model without lidar. Both with and without lidar, HCNAF outperforms a state-of-the-art prediction model, \textit{PRECOG-ESP} \cite{ref:2019_PRECOG}.}
    \label{Fig:Detailed_PRECOG_Evaluation}
\end{figure*}

\begin{figure*}[ht] 
    \centering
    \includegraphics[width=\linewidth]{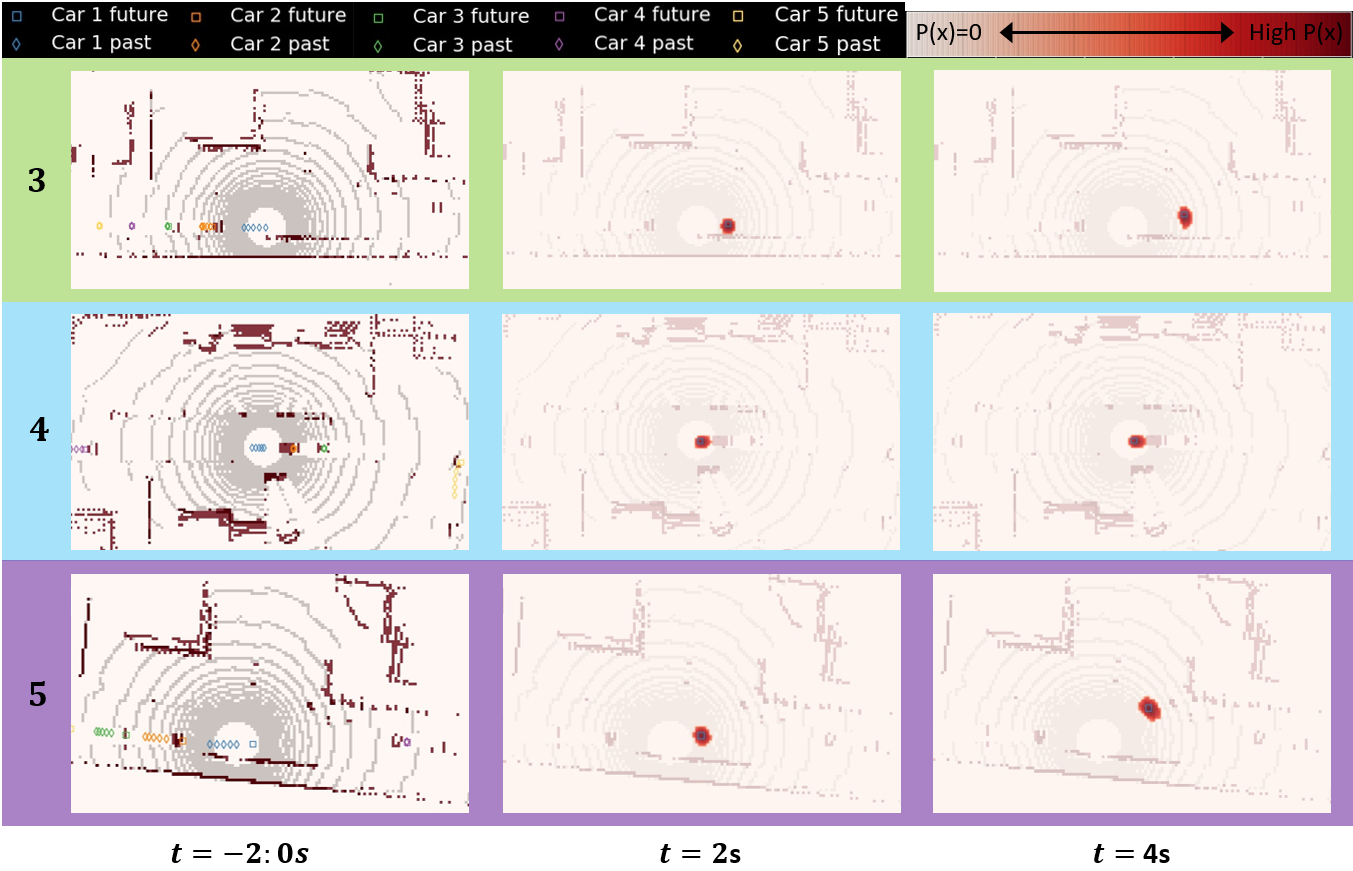}
    \caption{Continuing examples (3 through 5) from the POM forecasts of the HCNAF model described in Table \ref{Table:Exp_PRECOG_Carla} (with lidar) on the \textit{PRECOG-Carla} dataset. In the third example, the car 1 enters a 3-way intersection and our forecasts captures the two natural options (left-turn \& straight). Example 4 depicts a 3-way intersection with a queue formed by two other cars in front of car 1. HCNAF uses the interactions coming from the front cars and correctly forecast that car 1 is likely to stop due to other vehicles in front if it. In addition, our model captures possibilities of the queue resolved at $t$ = 4s and accordingly predicts occupancy at the tail. The fifth example illustrates car 1 while starting a turn left as it enters the 3-way intersection. The POM forecast for $t$ = 4s is an ellipse with a longer lateral axis, which reflects the higher uncertainty in the later position of the car 1 after the turning.}
    \label{Fig:PRECOG_POM1}
\end{figure*}

\begin{figure*}[ht] 
    \centering
    \includegraphics[width=\linewidth]{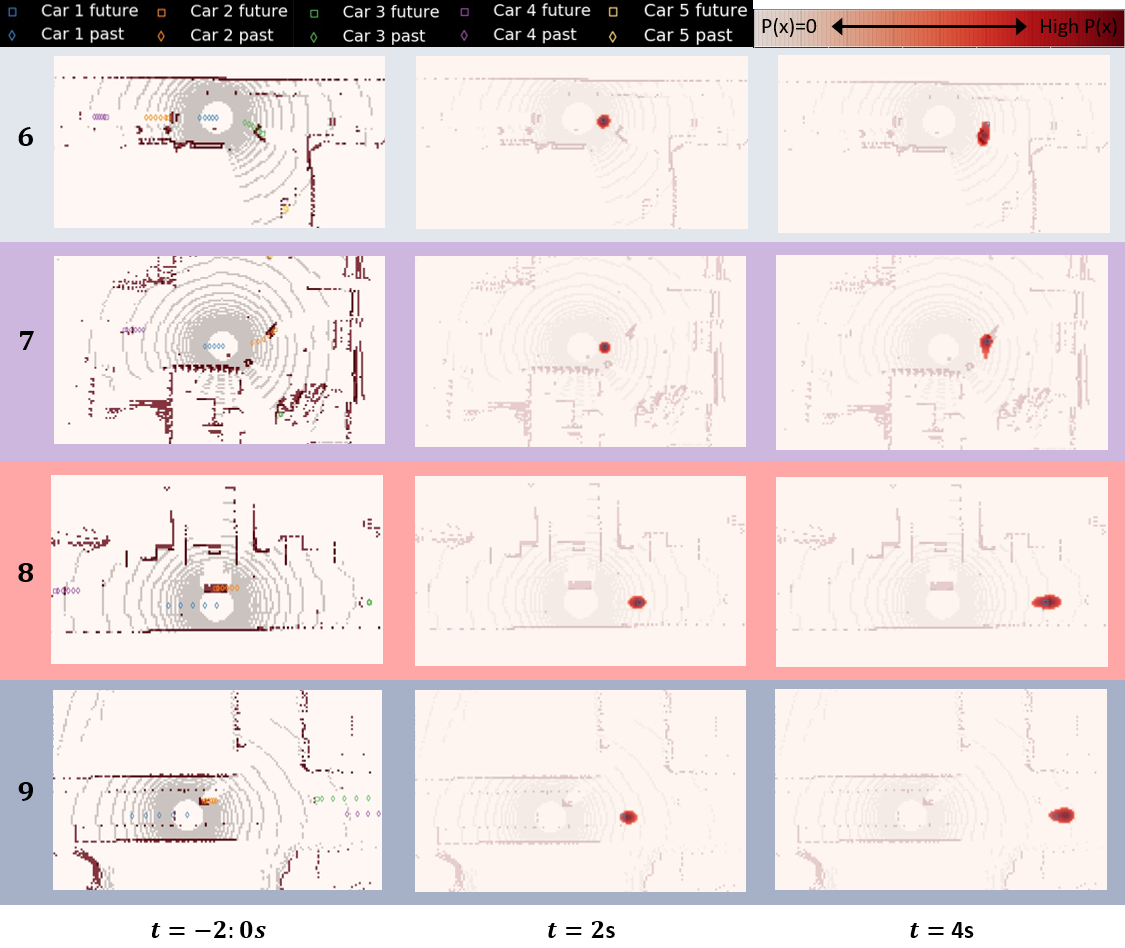}
    \caption{Continuing examples (6 through 9) from the POM forecasts of the HCNAF model described in Table \ref{Table:Exp_PRECOG_Carla} (with lidar) on the \textit{PRECOG-Carla} dataset. In examples 6 and 7, the car 1 enters 3-way intersections and POM shows that HCNAF forecast the multi-modal distribution successfully (straight \& right-turn for the example 6, left-turn \& right-turn for the example 7). Example 8 depicts a car traveling in high-speed in a stretch of road. The POM forecasts are wider-spread along the longitudinal axis. Finally, example 9 shows a car entering a 4-way intersection at high-speed. HCNAF takes into account the fact that car 1 has been traveling at high-speed and predicts the low likelihood of turning left or right, instead forecasting car 1 to proceed straight through the intersection.}
    \label{Fig:PRECOG_POM2}
\end{figure*}

\begin{figure*}[ht] 
    \centering
    \includegraphics[width=0.97\linewidth]{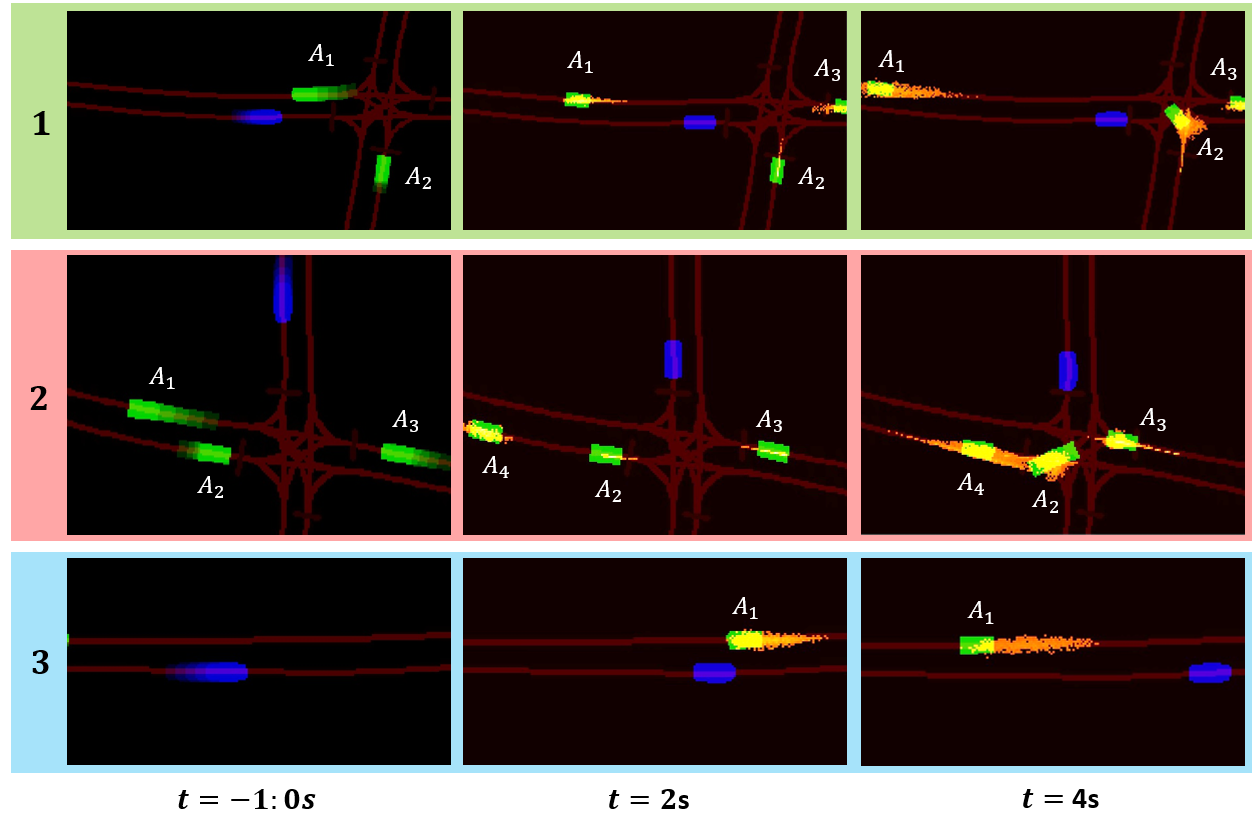}
    \caption{HCNAF for forecasting POMs on our \textit{Virtual Simulator} dataset. Left column: one-second history of actors (green) and reference car (blue). Actors are labeled as $A_{i}$. Center and right columns: occupancy prediction for actor centers $x_t$, $y_t$, at $t$ = 2 and 4 secs., with actor full body ground truth overlayed. Note that actors may enter and exit the scene. In example 1, our forecasts captured the speed variations of $A_1$, the stop line deceleration and the multi-modal movements (left/right turns, straight) of $A_2$, and finally the stop line pausing of $A_3$. In Example 2, HCNAF predicts $A_2$ coming to a stop and exiting the intersection before $A_{3}$, while $A_3$ is yielding to $A_2$. Finally, example 3 shows that HCNAF predicts the speed variations along a stretch of road for $A_1$.}
    \label{Fig:Vsim_POM}
\end{figure*}

\end{document}